# PHONOLOGY RECOGNITION IN AMERICAN SIGN LANGUAGE

*Federico Tavella*   *Aphrodite Galata*   *Angelo Cangelosi*

Department of Computer Science, The University of Manchester
{federico.tavella, a.galata, angelo.cangelosi}@manchester.ac.uk

## ABSTRACT

Inspired by recent developments in natural language processing, we propose a novel approach to sign language processing based on phonological properties validated by American Sign Language users. By taking advantage of datasets composed of phonological data and people speaking sign language, we use a pretrained deep model based on mesh reconstruction to extract the 3D coordinates of the signers keypoints. Then, we train standard statistical and deep machine learning models in order to assign phonological classes to each temporal sequence of coordinates.

Our paper introduces the idea of exploiting the phonological properties manually assigned by sign language users to classify videos of people performing signs by regressing a 3D mesh. We establish a new baseline for this problem based on the statistical distribution of 725 different signs. Our best-performing models achieve a micro-averaged F1-score of 58% for the major location class and 70% for the sign type using statistical and deep learning algorithms, compared to their corresponding baselines of 35% and 39%.

## 1. INTRODUCTION

Sign language (SL) is a natural language which uses visual and motor elements to deliver a message. As with the other forms of natural languages, sign language contains all fundamental features of a language, like word ordering/formation and pronunciation. Nevertheless, most of the computational research in sign language has focused on Sign Language Recognition (SLR) of single letters - i.e., fingerspelling - or words from images or videos [1, 2, 3, 4], sentence level recognition [5, 6] translation from sign to spoken language [7] and vice-versa [8]. This led to online tools based on machine learning that enable people to learn fingerspelling the alphabet using a webcam[1] or to detect sign language in online videoconferences [9]. Thus, SLR tools can significantly improve for SL users who communicate remotely with non sign language users.

[1] https://fingerspelling.xyz/

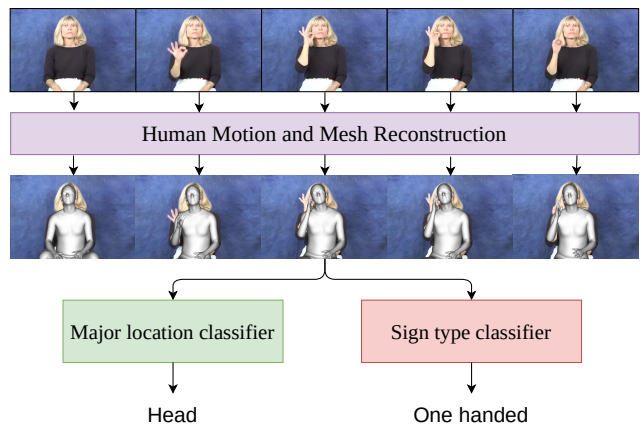

**Fig. 1**. Our approach extracts a mesh (and 3D coordinates) from videos of people speaking ASL and uses the keypoints to classify the video according to 3 phonological classes. For example, the sign "cat" is composed of a back and forth movement executed with one hand near the head.

Very few works use the **phonological properties** of sign language to perform SLR, most of them focusing on the hands pose. Cooper et al [10] perform sign language recognition based on lexical sub-units using a two stage approach. They extract 2D and 3D features using a Kinect sensor and train Hidden Markov Models (HMMs) and Sequential Pattern Boosting (SPB) classifiers to recognise a sign based on the subunits. To do so, the authors build an ad hoc dataset, which contains 984 Greek Sign Language signs with 5 examples of each performed by a single signer. Metaxas et al [11] use Conditional Random Fields (CRFs) to perform SLR, by creating a feature vector composed of 2D and 3D features extracted from videos via Convolutional Pose Machines [12] and additional linguistically relevant information. Camgoz et al [13] train a combination of CNN, LSTM and Connectionist Temporal Classification (CTC) in an end-to-end fashion to recognise phonologically meaningful properties to perform SL handshape recognition. Similarly, Borg et al [14] use a combination of tracking algorithms and RNN to perform word-level recognition based on subunits of hands pose and motion. Interestingly, they point out how the number of subunits in sign language is much smaller than the number of words. Thus, performing sign language recognition based on subunits instead of words is potentially much more efficient in terms of training data and number of possible outcomes, similarly to what is done in Optical Character Recognition (OCR), where models are trained on letters instead of words. For example, training a classifier to distinguish 10 different signs can be done with two different phonological classes, where class A has 2 possible values and class B has 5. This would require 7 examples instead of 10. None of the aforementioned studies take advantage of existing datasets containing phonological properties of sign language that were manually annotated by SL users.

We propose a mesh regression approach [15] to extract 3D temporal features from videos of people speaking American Sign Language (ASL) and a statistical [16] and deep learning [17] approach to perform a preliminary recognition of phonological classes based on the extracted features and ground truth assigned by ASL speakers. Figure 1 illustrates an example of our approach. We choose to infer phonological classes instead of words for the following reasons: (i) phonological properties of sign language are severely understudied in computational linguistic; (ii) training is more efficient in terms of resources; (iii) it improves the interpretability of the SRL models by providing specific features for each sign; (iv) being able to recognise phonological classes based on body movements can help to develop automated tools to teach people sign language by receiving automated feedback and improve tools for sign language generation; (v) developing a new tool could help linguists to perform new studies on sign language.

## 2. APPROACH

Our final aim is to be able to assign phonological classes to each of video. We divide our approach in different stages: data preparation, feature extraction, and model selection. We describe the available data and how we assign the labels for classification, illustrate the process used to extract 3D temporal features from the videos and define our models used for the classification of temporal sequences.

WLASL [18] is one of the largest datasets available composed of people demonstrating words in ASL. Each video has a *lemma* (i.e., dictionary form of a word) assigned as a label. However, we want to be able to identify phonological classes (e.g., the location where the sign is executed and the number of hands used to sign). Diversely, ASL-Lex [19] is a database of lexical and phonological properties of American Sign Language obtained through a study involving both deaf and hearing participants who were asked to assign some properties to videos of people performing signs. For each of the words and/or lemmas in the database, several different features are available: videoclip duration, sign length, grammatical class and phonological properties, such as sign type, selected fingers, flexion, major/minor location, or movement. A more detailed description can be found in [19] or using an online tool[2] which shows sign similarity using a graph. As a preliminary study in this direction, we are interested in phonological classes that can be represented using the full body. As such, we picked the following 2 classes: sign type (one vs two-handed, symmetrical or asymmetrical) and major location (spatial execution of the sign). Figure 2 presents the number of lemmas and possible values for the selected classes. Combining information from WLASL and ASL-Lex, our dataset is composed of a set of lemmas and their phonological properties obtained by cross-referencing the two datasets. It is composed of 790 videos and 993 lemmas with phonological properties. By matching the lemmas with the labels of the videos, we obtain our final dataset composed of 725 unique entries. The fact that there are no duplicate lemmas in our dataset means that there are no videos representing the same sign. Consequently, there are no two identical 3D sequences in the training, validation and test sets, reducing the possibility of introducing a bias in our evaluation.

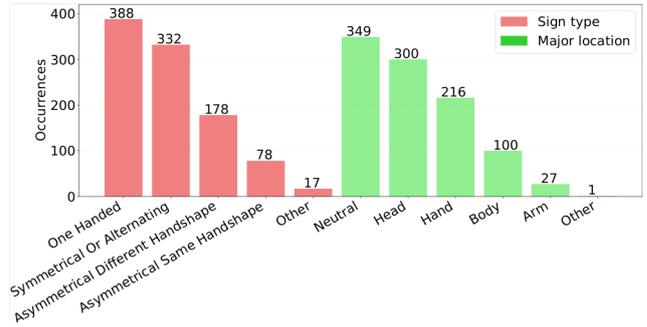

**Fig. 2.** Number of samples for each phonological class in ASL-Lex.

In order to extract 3D keypoints, we use the Human Mesh and Motion Recovery (HMMR) algorithm [15]. Given a video as input, HMMR extracts per-image features using a ResNet [20] network. Then, the features are used to train a temporal encoder so that it learns a representation of the 3D human dynamics over a temporal window centered at the current frame. Then, we can regress the 3D human shape and pose based on the Skinned Multi-Person Linear (SMPL) model [21]. Using a model such as SMPL brings the advantage of inferring parameters representing the human body, instead of directly regressing keypoints. Additionally, this allows us to compare different algorithms that use the same output model. Given the lack of 3D annotations regarding human poses, especially for videos, the HMMR is trained using heterogeneous datasets not related to ASL [22, 23, 24, 15]. For our purposes, we use HMMR in order to extract 3D coordinates for joints of the upper part of the body, namely head, chest, shoulders, elbows and wrists. For example, Figure 3 shows the projected 3D coordinates with respect to time of the sign "house", obtained via the HMMR model.

---
[2]https://ben-tanen.com/asl-lex/visualization/

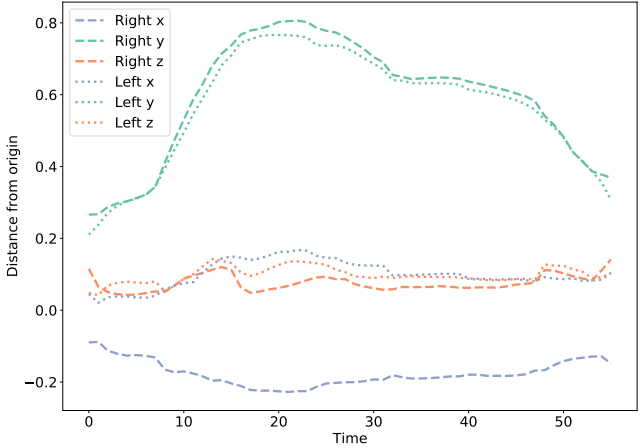

**Fig. 3**. Coordinates projection of the left and right wrists during the execution of the sign "house". While movements along the Y and Z axis are similar, there is a constant offset for the X axis. This indicates a symmetrical sign, which can be visualised using ASL-Lex online tool.

While assigning phonological labels can be - and has been - done manually, an automated approach would be much more efficient, removing the labour intensive manual labelling from researchers, who would just need to validate the data. As a baseline model, we use a classifier based on simple rules (e.g., predict the majority class or according to class distribution). Once we establish a baseline, we can move on to more complex models. Recurrent Neural Networks [25] have been broadly used for sequence classification and achieved promising results, but they usually require datasets much larger than ours to be trained. We use logistic regression and support vector machines as standard statistical models [16], multilayer perceptron as deep model [17] and Long Short-Term Memory (LSTM) [26] and Gated Recurrent Units (GRU) [27] as models capable of capturing the temporal component of sequences.

## 3. EXPERIMENTS

We test two different approaches, namely statistical models and deep learning, to infer the phonological classes associated to each video. We deploy a zero-padding strategy in order to make sure that all time series have the same duration and there is no loss of information (i.e., when compared to sequences subsampling). We then divide our data in a stratified (i.e., according to the labels distribution) fashion in two different subsets: training and test, with 85% of the data dedicated to the former and 15% to the latter. We use a 5-fold stratified cross-validation for hyperparameter tuning and we identify the micro-averaged F1-score as the best metric to train and measure the performance of our classifier, as the classes distribution of our dataset is uneven. As a first approach, we want to maximise the number of correct predictions, regardless of the dataset distribution. Micro-averaged F1-score gives us an idea of how well the classifier is performing *overall*, by calculating the total number of correctly classified instances instead of averaging the scores of the different classes. Morover, we report the macro-averaged F1-score because it is indicator about how the score is affected by the imbalance across different classes. Each of the final evaluations runs is repeated 10 times using different seeds to calculate the mean and standard deviation.

Multinomial logistic regression is the first algorithm that we are exploring for classification. A simple method is ideal as a first approach to understand how much can be learned from the data as there is a small number of parameters that we can modify. In particular, we investigate different techniques of regularization in order to prevent overfitting. Support Vector Machines (SVMs) [28] are a powerful tool to perform non-linear classification. Their power relies on the flexibility they provide using the *kernel function*, which makes them operate efficiently in high-dimensional space. Thus, the kernel function is certainly among the most important parameters to select for a good performance. We want our models to perform better than the baselines, but at the same time they should be flexible enough to generalise. For logistic regression and SVM, this means considering the number of iterations used to train the models and the regularisation magnitude. In order to take advantage of the temporal component of our data, we can use deep architectures that are explicitly designed to do so, such as RNN [25]. In particular, we are going to use LSTM [26] and GRU [27]. When using neural networks, each single parameter and hyperparameter can make impact on the model performance. Thus, we focus our experiments on the neural network architecture and methods to avoid overfitting, such as dropout and batch normalization, given the small size of our dataset. We also train a multi-layer perceptron as a middle ground between statistical and deep learning methods.

## 4. RESULTS AND DISCUSSION

We provide the results of the set of hyperparameters leading to the highest score. In order to calculate a baseline for our current set-up, we take into account the metrics used to measure the performance and the data distribution. Given that we chose the micro-averaged F1-score, the baseline which yields the highest score is the one that generates predictions based on the majority class. In addition, we provide a baseline and results for the macro-averaged F1-score as an additional evaluation metric. The macro-averaged baseline is calculated using the statistical distribution (i.e., the number of times a certain value appears compared to the total number of samples) of the different labels. For the major location, the micro and macro F1 scores are respectively 34.9% and 20.7%, while for the sign type are 38.5% and 20.0%.

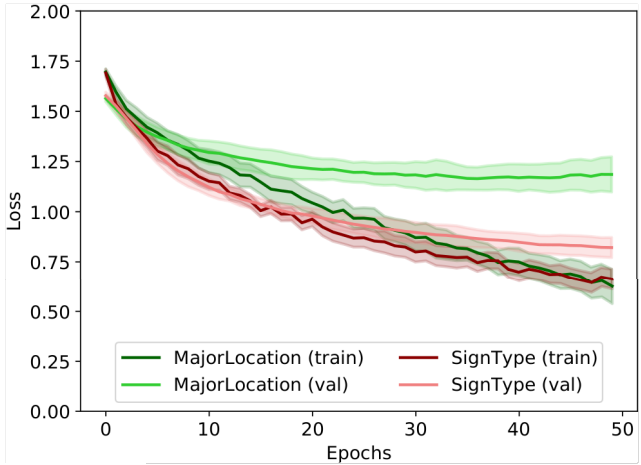

**Fig. 4**. MLP learning curves (train and validation) for "major location" and "sign type" features.

We train our 3 different models for 50 epochs using the Adam [29] optimisation algorithm and a learning rate equal to $10^{-4}$. Figure 4 illustrates the learning curves (i.e., training and validation loss) of our best performing MLP model. We found this configuration to be a good trade-off between training speed and fluctuations in the loss. Based on the validation loss, we can see that 50 epochs are enough to make the models converge, despite the limited amount of data.

Table 1 summarises the results for each different model and compares them with the baseline. Overall, most of the models perform similarly over the same class, with neural networks-based models that usually achieve slightly higher scores than standard models. For the sign type, statistical models achieve around 53% of correct predictions, while deep learning models measure around 57% on average, both accounting for an increment around 55% when compared to the baseline. For the major location, there is no relevant difference between statistical and deep models, with both scoring approximately 68% for the micro F1 score, an increment of 74% compared to the baseline. Finally, all the models experience a 100% gain compared to the baseline when considering the macro F1 score, indicating that not only the models learn to predict the classes that are most represented, but also according to the dataset distribution despite being trained to maximise the number of correct predictions.

Figure 5 illustrates the confusion matrix for a single run of the MLP trained to recognise the sign type. We can see that the algorithm mainly distinguishes one vs two handed signs. Moreover, removing the strongly underrepresented value "Other" would lead to an increase in the actual score of the model. Lastly, asymmetrical signs with different and same hand shapes (ADH and ASH) are classified as symmetrical or alternating (SOA) due to the lack of features regarding

|  | Major location | | Sign type | |
| --- | --- | --- | --- | --- |
| Model | micro F1 | macro F1 | micro F1 | macro F1 |
| Baseline | 34.9 ± 0.0 | 20.7 ± 3.1 | 38.5 ± 0.0 | 20.0 ± 3.0 |
| Logistic | 53.2 ± 2.5 | 43.1 ± 7.4 | 67.2 ± 2.6 | 42.5 ± 4.2 |
| SVM | 52.8 ± 3.3 | 43.9 ± 5.9 | 68.6 ± 1.9 | 40.5 ± 2.5 |
| MLP | 58.1 ± 3.6 | 42.7 ± 3.8 | 69.1 ± 4.2 | 41.8 ± 4.3 |
| LSTM | 56.8 ± 2.9 | 42.8 ± 4.1 | 70.2 ± 3.5 | 42.2 ± 2.5 |
| GRU | 56.1 ± 3.9 | 43.2 ± 5.3 | 69.4 ± 3.8 | 42.7 ± 2.4 |

**Table 1**. Summary of the results for each model trained with the hyperparameters that respectively lead to the best micro-averaged F1 score. Test scores calculated over 10 different seeds.

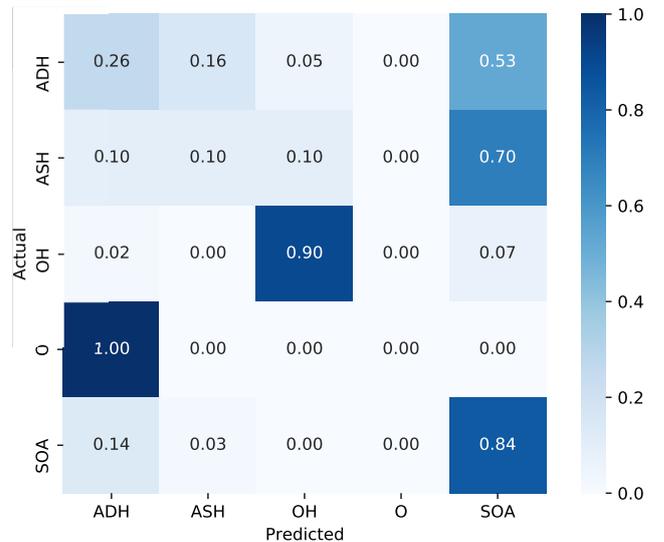

**Fig. 5**. Confusion matrix of MLP for the "sign type" feature. Labels for each row/column are assigned using the first letters of the phonological classes.

the hand pose.

## 5. CONCLUSION

In this paper, we propose a novel approach to the recognition of phonological classes of ASL based on data validated by ASL users. We extract temporal features from the video using a 3D tracking algorithm based on the SMPL model and perform recognition using statistical and deep models. Our experiments suggest that it is possible to extract phonological classes from videos using pre-trained tracking algorithms. We believe this work opens many possibilities for additional research. In our future work, we are planning to address our approach shortcomings by expanding the dataset, including data augmentation techniques to compensate imbalance and replace the current tracking algorithm with one that includes hands and facial features.